\pdfoutput=1

\documentclass[11pt]{article}

\usepackage{acl}

\usepackage{times}
\usepackage{latexsym}

\usepackage{graphicx}
\usepackage{subcaption,booktabs}
\usepackage{tabularx}
\usepackage{multirow, amsmath}
\usepackage{placeins}
\usepackage{gensymb} 
\usepackage{amssymb}

\usepackage{amsthm}
\usepackage{color, colortbl}
\usepackage{float}
\usepackage{enumitem}
\usepackage{bbm}

\usepackage[T1]{fontenc}

\usepackage[utf8]{inputenc}

\usepackage{microtype}

\usepackage{inconsolata}

\title{\includegraphics[width=1.2em, trim=25 75 25 25, clip]{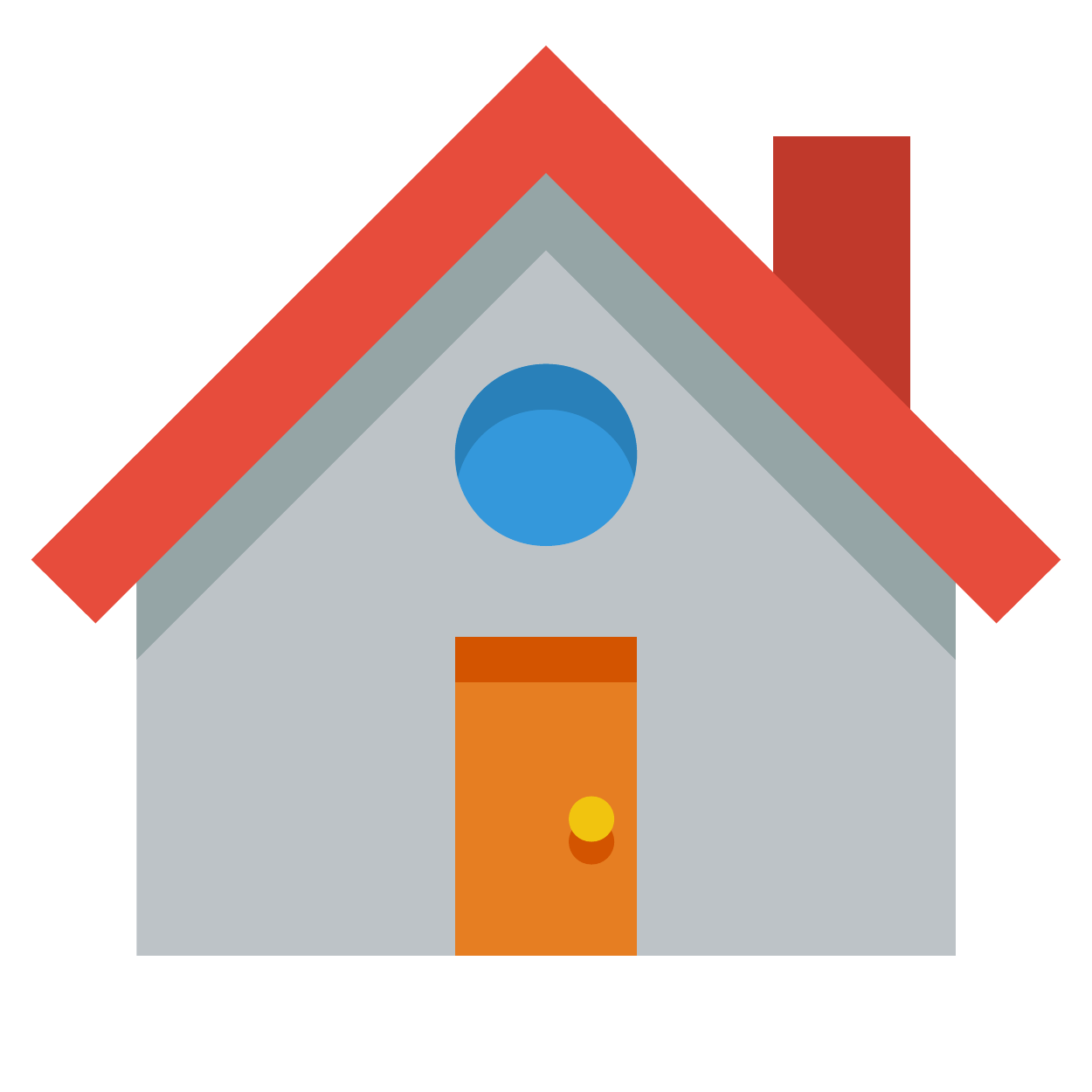} \textsc{Casa}: Causality-driven Argument Sufficiency Assessment}

\author{
	Xiao Liu$^{1}$\thanks{\;\;Work done during visiting UCLA.},\;
	Yansong Feng$^{1}$\and
	Kai-Wei Chang$^{2}$ \\
    $^1$Wangxuan Institute of Computer Technology, Peking University\\
	$^2$Computer Science Department, University of California, Los Angeles\\
	{\tt \{lxlisa,fengyansong\}@pku.edu.cn} \\
	{\tt kwchang@cs.ucla.edu}\\
}

\begin{document}
\maketitle
\begin{abstract} 
The argument sufficiency assessment task aims to determine if the premises of a given argument support its conclusion.
To tackle this task, existing works often train a classifier on data annotated by humans. However, annotating data is laborious, and annotations are often inconsistent due to subjective criteria. Motivated by the definition of probability of sufficiency (PS) in the causal literature, we propose \textsc{Casa}, a zero-shot causality-driven argument sufficiency assessment framework. 
PS measures how likely introducing the premise event would lead to the conclusion when both the premise and conclusion events are absent. To estimate this probability, we propose to use large language models (LLMs) to generate contexts that are inconsistent with the premise and conclusion and revise them by injecting the premise event.
Experiments on two logical fallacy detection datasets demonstrate that \textsc{Casa} accurately identifies insufficient arguments. We further deploy \textsc{Casa} in a writing assistance application, and find that suggestions generated by \textsc{Casa} enhance the sufficiency of student-written arguments. Code and data are available at \url{https://github.com/xxxiaol/CASA}.
\end{abstract}

\section{Introduction}
\label{sec:intro}
\begin{figure}[t]
    \centering
    \includegraphics[width=0.9\columnwidth]{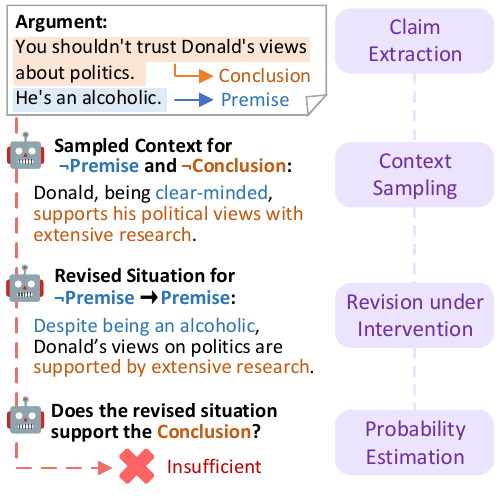}
    \caption{An example of the argument sufficiency assessment task and the reasoning steps of \textsc{Casa}.}
    \label{fig-intro}
\end{figure}

Argumentation is an integral part of our daily verbal communication~\citep{fogelin2005understanding, stab-gurevych-2017-parsing}. An argument is a series of statements consisting of premises and a conclusion. Take the argument shown in Figure~\ref{fig-intro} as an example: \textit{You shouldn't trust Donald's views about politics. He's an alcoholic.} The first sentence, which serves as the conclusion, is supported by the second sentence, acting as the premise. If we have techniques to assess the quality of arguments precisely, we can identify weaknesses in arguments and further improve them.

An important part of argument quality assessment is to determine whether the premises sufficiently support the conclusion. In a cogent argument, the premises are not only relevant to the conclusion and acceptable on their own, but also collectively sufficient to draw the conclusion~\citep{blair2011groundwork}. 
We focus on the sufficiency assessment in this paper. In the example of Figure~\ref{fig-intro}, the premise \textit{Donald is an alcoholic} does not sufficiently support the conclusion \textit{his views are untrustworthy}, as there are other factors that could invalidate the conclusion. For instance, if \textit{Donald's views on politics are supported by extensive research}, his views could still be credible even though he is an alcoholic. 

Previous works on argument sufficiency assessment train classifiers based on human annotations~\citep{gurcke-etal-2021-assessing, saveleva-etal-2021-graph}. However, the sufficiency criteria are vague and subjective among annotators. For example,~\citet{wachsmuth-etal-2017-computational} collected annotations from seven annotators, but even the three annotators with the highest consensus only achieved an agreement of 0.28.\footnote{The agreement is measured with Krippendorff's $\alpha$. $\alpha= 0.67$ is suggested as the lowest acceptable limit for tentative conclusions~\citep{krippendorff2018content}.} This inconsistency poses a challenge in learning an accurate model. 

In this paper, we propose \textsc{Casa}, a zero-shot \textbf{C}ausality-driven \textbf{A}rgument \textbf{S}ufficiency \textbf{A}ssessment framework by formulating the task with a concept borrowed from causality: the Probability of Sufficiency (PS)~\citep{pearl2000models}.
Intuitively, if $X$ is a sufficient cause of $Y$, the presence of $X$ implies the subsequent occurrence of $Y$.
PS quantifies the probability that introducing $X$ would produce $Y$ in the case where $X$ and $Y$ are in fact absent:
\begin{equation*}
    PS_{X, Y} = P(Y(X=1)=1|X=0, Y=0),
\end{equation*}
where $Y(X=1)$ indicates the value of $Y$ after an intervention on $X$.
Take the example in Figure~\ref{fig-intro}. $X$ is the occurrence of the event \textit{Donald is an alcoholic}, and $Y$ represents \textit{Donald's views about politics are untrustworthy}. If $X$ and $Y$ are both false, but when the event \textit{Donald becomes an alcoholic} occurs, it results in \textit{ his political view being untrustworthy}, then the argument is sufficient. 


To measure PS of a given argument, there presents the following challenges:
1) How to measure the probabilities without observational data, i.e., how to estimate $P(Y=1|X=0, Y=0)$ if we do not have the corresponding data points. 2) Even if we have the observational data, how to intervene in the argument, i.e., how to estimate $P(Y(X=1)=1)$ given data conforming to the conditions of $X=0$ and $Y=0$.

Our approach tackles the challenges by leveraging the commonsense knowledge and reasoning abilities of large language models (LLMs)~\citep{bhargava2022commonsense, kojima2022large}.
Specifically, we ask LLMs to sample data that are inconsistent with the premises and the conclusion, such as \textit{ Donald, being clear-minded, supports his political views with extensive research}; and then revise the data to contain the premises, such as \textit{ despite being an alcoholic, Donald’s views on politics are supported by extensive research}. Step-wise evaluation results demonstrate the effectiveness of using LLMs to sample data and to conduct interventions.

We evaluate \textsc{Casa}'s capability of assessing argument sufficiency on two logical fallacy detection datasets, BIG-bench Logical Fallacy Detection~\citep{srivastava2023beyond} and Climate~\citep{alhindi-etal-2022-multitask}.
We compare our framework with baseline methods including zero-shot/one-shot prompting and perplexity-based classification, and find that \textsc{Casa} distinguishes between sufficient and insufficient arguments more accurately, bringing an average of 10\% improvement than directly prompting the same base models.

To further investigate whether our framework can help in realistic scenarios, we apply \textsc{Casa} to provide writing suggestions for student essays. \textsc{Casa} generates \textit{objections} (reasons arguing against the argument) for arguments it finds insufficient in essays. We conduct a human evaluation to assess the quality of \textsc{Casa}'s suggestions and their effects on revision. Results demonstrate that the objections generated by \textsc{Casa} are rational and help improve the sufficiency of the arguments.

Our main contributions are as follows:
1) We design \textsc{Casa}, a theoretically grounded framework for argument sufficiency assessment based on the probability of sufficiency.
2) To realize the probability of sufficiency, we exploit LLMs in generating data samples and conducting interventions, and demonstrate the effectiveness with experiments on logical fallacy detection.
3) We demonstrate a practical application of \textsc{Casa} in improving the sufficiency of arguments in student-written essays.

\section{The \textsc{Casa} Framework}
\begin{figure*}[th]
    \centering
    \includegraphics[width=\textwidth]{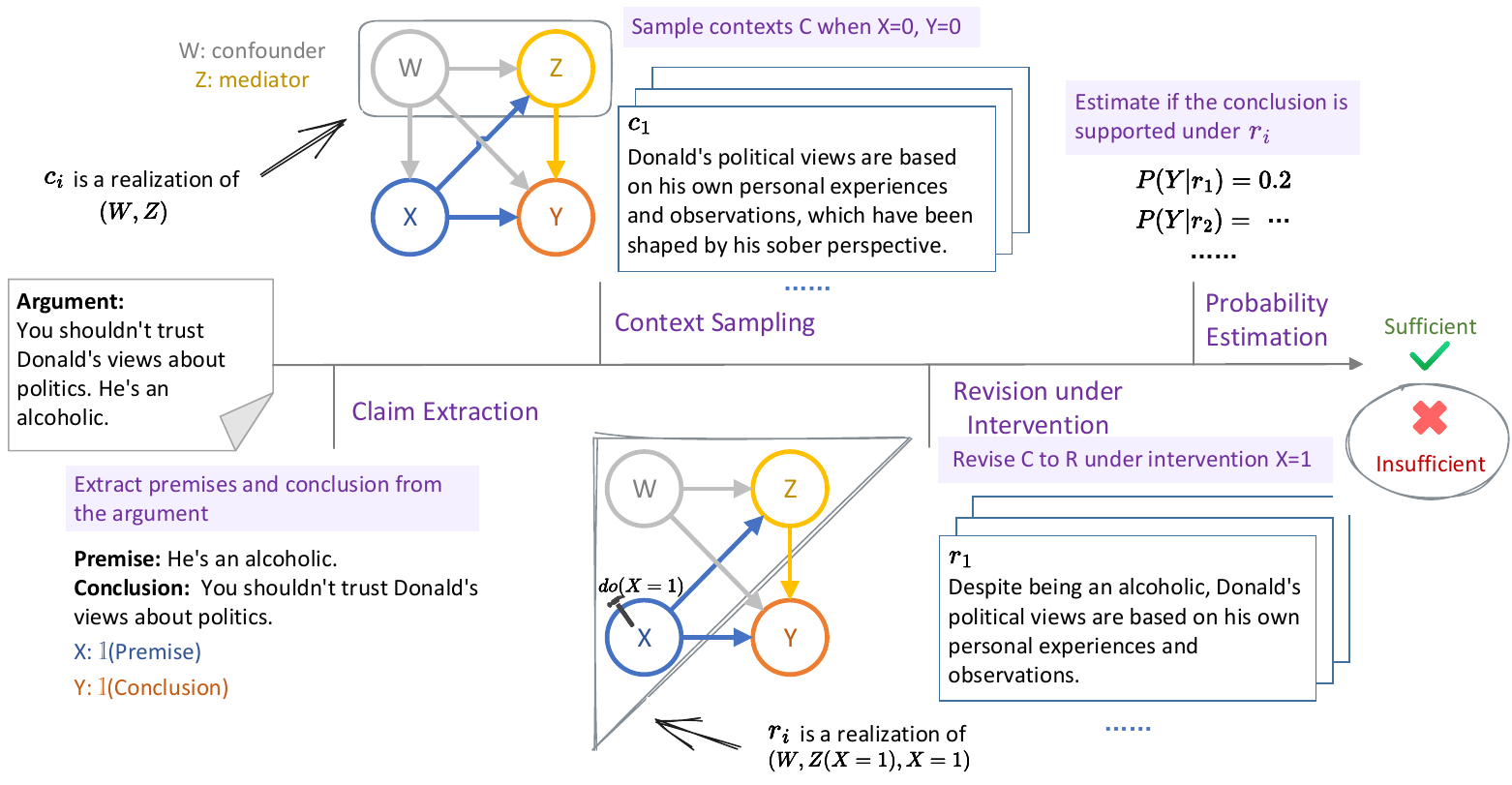}
    \caption{Overall architecture of \textsc{Casa}. Given an argument, we first extract its premises and conclusion, sample contexts that are inconsistent with the premises and conclusion, revise the contexts to meet the premises, and finally estimate the probability of the conclusion. Edge $A \rightarrow B$ in causal graphs means $A$ may causally affects $B$.} 
    \label{fig-method}
\end{figure*}

\noindent\textbf{Notations.}
We define $X$ as the occurrence of a premise event\footnote{For simplicity, we consider only arguments with a single premise first. We will discuss how to extend the discussion to arguments with multiple premises in Section \ref{sec:multipremises}.}: $X=\mathbbm{1}(\texttt{Premise})$, and $Y$ as the occurrence of a conclusion event: $Y=\mathbbm{1}(\texttt{Conclusion})$. Here $\mathbbm{1}(\cdot)$ is the indicator function. Both $X$ and $Y$ are binary variables. $Y_u$ indicates the value of $Y$ in the unit $u$.

\vspace{1mm}
\noindent\textbf{Assumptions.}
Our framework is based on two common assumptions in causal inference~\citep{rubin1978bayesian}:
\begin{itemize}[noitemsep, nolistsep]
    \item \textit{No interference}: the value $Y$ of the unit $u$ is not affected by the values of $X$ assigned to other units. 
    \item \textit{Consistency}: $X=x \rightarrow Y=Y(X=x)$, where $x$ indicates a specific value of $X$. This requires that each treatment value $x$ has only one form~\citep{imbens2015causal}.
\end{itemize}
In our task, the first assumption is satisfied as there is no dependency between the conclusion of one unit and the premise of another. To satisfy the second assumption, we restrict the value $X=0$ to the occurrence of $\neg \texttt{Premise}$, the negated form of the premise event. For coherence, $Y=0$ indicates the occurrence of $\neg \texttt{Conclusion}$.

\vspace{1mm}
\noindent\textbf{Overview.}
As shown in Figure~\ref{fig-method}, we first extract the premise and conclusion from a given argument, then sample contexts that meet the conditions, make interventions on the contexts, and finally estimate the probability of the conclusion for each unit.

\subsection{Claim Extraction}
Given an unstructured argument, we aim to split it into multiple premises and one conclusion. The task of argument parsing is indeed complex~\citep{ajjour-etal-2017-unit},\footnote{Previous works either use pre-extracted premises and conclusion~\citep{gurcke-etal-2021-assessing}, or train an individual segmentation model based on annotations~\citep{saveleva-etal-2021-graph}.} but since it is not the primary component of our framework, we simplify it in two ways: 
1) We segment the argument into claims with punctuation marks and conjunction words; 2) we do not consider how the premises are related to each other, such as how one premise might support another premise.
Specifically, we list the segmented claims, ask an LLM to select which one is the conclusion, and consider the other claims as premises. 

\subsection{Context Sampling}
To calculate the conditional probability, an intuitive way is to sample $n$ units $U=\{u_1, ..., u_n\}$ that conform to the conditions $X=0$ and $Y=0$. Although we do not have existing observational data, we make use of the commonsense knowledge learned by LLMs, and let them generate $n$ diverse contexts $C=\{c_1, ..., c_n\}$ which are consistent with $\neg \texttt{Premise}$ and $\neg \texttt{Conclusion}$. 

Take the argument in Figure~\ref{fig-method} as an example. We instruct the LLM to ``generate $n$ detailed contexts. Each context is consistent with both the premise \textit{Donald isn't an alcoholic} and the conclusion \textit{you should trust Donald's views about politics.}'' A generated context mentions that \textit{Donald's political views are based on his own personal experiences
and observation}, which is consistent with $\neg \texttt{Conclusion}$; and his views \textit{have been shaped by his sober perspective}, consistent with $\neg \texttt{Premise}$.

$PS_{X, Y}$ is then estimated with the average of $P(Y_{u_i}(X=1)=1)$ under each unit $u_i$:
\begin{equation*}
    PS_{X, Y} \approx \frac{1}{|U|}\sum_{u_i\in U|X=0, Y=0}\!\! P(Y_{u_i}(X=1)=1).
\end{equation*}
From the causal lens, we can decompose the context information into two latent parts $W$ and $Z$. $W$ is the part that is not causally affected by $X$, called confounder; and $Z$ is the remaining part that may be causally affected by $X$, called mediator.\footnote{Our naming conforms to the common terminology. As shown in the upper left of Figure~\ref{fig-method}, $W$ is called confounder because there may be causal relations $W\rightarrow X$ and $W\rightarrow Y$, and $Z$ is called mediator because there may be causal relations $X\rightarrow Z$ and $Z\rightarrow Y$.} Each context $c_i$ can be seen as a realization of $(W, Z)$.

\subsection{Revision under Intervention}
For each unit $u_i=(c_i, X=0, Y=0)$, our next step is to implement the intervention $X=1$ on it. 
The effect of the intervention on the context $c_i$ is illustrated by the causal graph  
in the lower part of Figure~\ref{fig-method}. The intervention breaks the causal relation $W \rightarrow X$, and leaves $W$ unchanged. At the same time, $Z$ will change according to the intervention, becoming $Z(X=1)$. Therefore, we can rewrite the estimand as:
\begin{equation*}
\begin{aligned}
    P(Y_{u_i} & (X=1)=1)=\\
    & P(Y_{u_i}(X=1, Z(X=1))=1).
\end{aligned}
\end{equation*}
We ask the LLM to revise each context $c_i$ to $r_i$ under intervention $X=1$. Specifically, our instruction is to revise the context to contain the premise,\footnote{An example prompt is in Appendix Table~\ref{table-prompt-revision}.} so the information $X=1$ is also included in $r_i$, leading $r_i$ to be a realization of $(W, Z(X=1), X=1)$.

In Figure~\ref{fig-method}, the expression \textit{sober perspective} is removed by the LLM, and the $\texttt{Premise}$ is added with \textit{Despite being an alcoholic}. At the same time, \textit{Donald's political views are based on his own personal experiences
and observation} is kept unchanged, as this does not violate the $\texttt{Premise}$. 

\subsection{Probability Estimation}
We transform the probability estimation of $Y$ into the form of natural language inference (NLI): under each situation $r_i$, estimate whether the conclusion is supported or contradicted. We use an off-the-shelf NLI model to make the prediction, and aggregate the units to calculate the final $PS_{X, Y}$.

\subsection{Dealing with Multiple Premises}
\label{sec:multipremises}
When an argument contains multiple premises, we exhaustively check the sufficiency of each premise given the context of other premises:
\begin{equation*}
\begin{aligned}
PS_{X_i, Y|X_{1\cdots n\backslash i}}=&P(Y(X_i=1)=1|X_i=0,\\
&Y=0, X_{1\cdots n\backslash i}=\mathbf{1}).
\end{aligned}
\end{equation*}
Concretely, we ask the LLM to contain $\texttt{Premise}_{1\cdots n\backslash i}$ when sampling contexts for checking the sufficiency of the $i$-th premise event.
\section{Experiments}
\begin{table*}[ht]
    \centering
    \small
    \begin{subtable}[h]{0.48\textwidth}
        \centering
        \begin{tabular}{lcc}
        \toprule
        \textbf{Model} & \textbf{Acc} & \textbf{Macro-F1} \\
        \midrule
        \rowcolor[gray]{0.95} \textit{Unsupervised} &  &  \\
        Zero-shot Prompting (\textsc{Tulu}) & 59.5 & 59.3 \\
        Zero-shot Prompting (\textsc{Llama}2) & 70.0 & 66.8 \\
        Perplexity (\textsc{Tulu}) & 56.0 & 54.7 \\
        Perplexity (\textsc{Llama}2) & 51.0 & 50.8 \\
        DeBERTa-NLI & 57.0 & 54.1 \\
        BART-NLI & 67.0 & 65.5 \\
        \textsc{Casa} (\textsc{Tulu}) & 77.0 & 70.8 \\
        \textsc{Casa} (\textsc{Llama}2) & \textbf{79.0} & \textbf{73.4}\\
        \arrayrulecolor{black!30}\midrule
        One-shot Prompting (\textsc{Tulu}) & 61.1 & 59.7 \\
        One-shot Prompting (\textsc{Llama}2) & 74.1 & 68.6 \\
        \arrayrulecolor{black}\bottomrule
        \end{tabular}
        \caption{Results on BIG-bench-LFD.}
    \end{subtable}
    \hspace{6pt}
    \begin{subtable}[h]{0.48\textwidth}
        \centering
        \begin{tabular}{lcc}
        \toprule
        \textbf{Model} & \textbf{Acc} & \textbf{Macro-F1} \\
        \midrule
        \rowcolor[gray]{0.95} \textit{Unsupervised} &  &  \\
        Zero-shot Prompting (\textsc{Tulu}) & 33.0 & 30.3 \\
        Zero-shot Prompting (\textsc{Llama}2) & 51.9 & 48.0 \\
        Perplexity (\textsc{Tulu}) & 63.2 & 38.7 \\
        Perplexity (\textsc{Llama}2) & 66.9 & 45.0 \\
        DeBERTa-NLI & 55.7 & 51.5 \\
        BART-NLI & 63.2 & 53.2 \\
        \textsc{Casa} (\textsc{Tulu}) & 64.2 & 54.9 \\
        \textsc{Casa} (\textsc{Llama}2) & \textbf{67.9} & \textbf{61.2}\\
        \arrayrulecolor{black!30}\midrule
        One-shot Prompting (\textsc{Tulu}) & 45.6 & 45.5 \\
        One-shot Prompting (\textsc{Llama}2) & 52.8 & 51.1 \\
        \arrayrulecolor{black}\bottomrule
        \end{tabular}
        \caption{Results on Climate.}
    \end{subtable}
    \caption{Automatic evaluation results of argument sufficiency assessment, showing that \textsc{Casa} outperforms all zero-shot and one-shot baselines. Numbers are in percentages (\%).}
    \label{table-main}
\end{table*}
\begin{table}[ht]
\centering
\small
\scalebox{0.85}{
\begin{tabular}{lcc}
    \toprule
     & \textbf{\textsc{Casa} (\textsc{Tulu})} & \textbf{\textsc{Casa} (\textsc{Llama2})} \\
    \midrule
    \rowcolor[gray]{0.95} \textit{Claim Extraction} & & \\
    \quad Correctness & 93\% & 92\% \\
    \rowcolor[gray]{0.95} \textit{Context Sampling} &  &  \\
    \quad Consistency with X=0 & 96\% & 90\% \\
    \quad Consistency with Y=0 & 91\% & 93\% \\
    \rowcolor[gray]{0.95} \textit{Revision under Intervention} &  &  \\
    \quad Consistency with X=1 & 95\% & 96\% \\
    \bottomrule
\end{tabular}
}
\caption{Step-wise evaluation on BIG-bench-LFD.}
\label{table-stepwise}
\end{table}
\subsection{Datasets}
We conduct experiments on two English logical fallacy detection datasets: BIG-bench Logical Fallacy Detection (BIG-bench-LFD)~\citep{srivastava2023beyond} and Climate~\citep{alhindi-etal-2022-multitask}. Logical fallacy detection requires models to distinguish between fallacious and correct arguments. As~\citet{jin-etal-2022-logical} mentioned, logical fallacies usually happen when the premises are insufficient to draw the conclusion. We manually check both datasets and confirm that the fallacious arguments can be attributed to insufficiency.

Due to the subjective criteria of argument sufficiency annotation mentioned in Section \ref{sec:intro}, the existing argument sufficiency datasets are noisy. We do not use them for automatic evaluation, but they will be used in the application of writing suggestions in Section~\ref{sec-application}. In contrast, logical fallacy datasets are more objective, with a clear distinction between fallacious and correct arguments.

\vspace{1mm}
\noindent\textbf{BIG-bench-LFD.} This dataset aims to evaluate LLMs' capabilities of detecting informal and formal fallacies. We only consider the informal statement portion, whose statements are more similar to real arguments. They are examples of good and bad cases of informal reasoning collected from philosophers, including 57 correct and 143 fallacious arguments.

\vspace{1mm}
\noindent\textbf{Climate.} This dataset contains arguments from climate change articles fact-checked by climate scientists at \url{climatefeedback.org}. Because some arguments in this dataset are single claims without premises, we only use instances with more than one sentence to avoid these single claim arguments, resulting in 30 correct and 76 fallacious arguments.

Accuracy and macro-F1 are reported for both datasets.

\begin{table*}[ht]
    \centering
    \small
    \begin{subtable}[h]{0.48\textwidth}
        \centering
        \scalebox{0.9}{
        \begin{tabular}{lcc}
        \toprule
        \textbf{Model} & \textbf{Acc} & \textbf{Macro-F1} \\
        \midrule
        \rowcolor[gray]{0.95} \textsc{Casa} (\textsc{Tulu}) & 77.0 & \textbf{70.8} \\
        \quad w/o Intervention & 74.5 & 62.5 \\
        \quad w/o Condition on $X=0$ & 75.0 & 68.7 \\
        \quad w/o Condition on $Y=0$ & \textbf{78.0} & 69.8 \\
        \quad Intervention: Concatenation & 74.0 & 63.3 \\
        \bottomrule
        \end{tabular}
        }
        \caption{Ablations for \textsc{Casa} (\textsc{Tulu}).}
    \end{subtable}
    \hspace{6pt}
    \begin{subtable}[h]{0.48\textwidth}
        \centering
        \scalebox{0.9}{
        \begin{tabular}{lcc}
        \toprule
        \textbf{Model} & \textbf{Acc} & \textbf{Macro-F1} \\
        \midrule
        \rowcolor[gray]{0.95} \textsc{Casa} (\textsc{Llama}2) & \textbf{79.0} & \textbf{73.4}\\
        \quad w/o Intervention & 73.5 & 65.8 \\
        \quad w/o Condition on $X=0$ & 75.5 & 67.1 \\
        \quad w/o Condition on $Y=0$ & 75.0 & 66.7 \\
        \quad Intervention: Concatenation & 78.0 & 67.4 \\
        \bottomrule
        \end{tabular}
        }
        \caption{Ablations for \textsc{Casa} (\textsc{Llama2}).}
    \end{subtable}
    \caption{Ablation results on BIG-bench-LFD. All ablations lead to performance drops on macro-F1. }
    \label{table-ablation-bigbench}
\end{table*}
\begin{table*}[ht]
    \centering
    \small
    \renewcommand{\arraystretch}{1.2}
    \begin{tabularx}{\textwidth}{XX}
    \toprule
    \multicolumn{2}{>{\hsize=\dimexpr2\hsize+2\tabcolsep+\arrayrulewidth\relax}X}{\textbf{Argument:} Biological, geological and planetary systems are extremely robust. Our evolving dynamic planet has survived sea level changes of hundreds of metres.}\\
    \textbf{Correct Answer:} Insufficient & \textbf{Zero-shot Prompting (\textsc{Tulu}):} Sufficient \\
    \rowcolor[gray]{0.95} \multicolumn{2}{>{\hsize=\dimexpr2\hsize+2\tabcolsep+\arrayrulewidth\relax}X}{\textsc{Casa} (\textsc{Tulu})}\\
    \multicolumn{2}{>{\hsize=\dimexpr2\hsize+2\tabcolsep+\arrayrulewidth\relax}X}{\textbf{Premise:} Our evolving dynamic planet has survived sea level changes of hundreds of metres.} \\
    \multicolumn{2}{>{\hsize=\dimexpr2\hsize+2\tabcolsep+\arrayrulewidth\relax}X}{\textbf{Conclusion:} Biological, geological and planetary systems are extremely robust.} \\
    \multicolumn{2}{>{\hsize=\dimexpr2\hsize+2\tabcolsep+\arrayrulewidth\relax}X}{\textbf{Sampled Contexts:}} \\
    \multicolumn{2}{>{\hsize=\dimexpr2\hsize+2\tabcolsep+\arrayrulewidth\relax}X}{\quad 1. The rapid rise of sea levels caused by climate change has led to the destruction of many coastal cities and ecosystems, demonstrating the vulnerability of biological, geological, and planetary systems.} \\
    \multicolumn{2}{>{\hsize=\dimexpr2\hsize+2\tabcolsep+\arrayrulewidth\relax}X}{\quad 2. The geological history of our planet is marked by numerous catastrophic events, such as massive volcanic eruptions and asteroid impacts, which have had a significant impact on the evolution of life on Earth.} \\
    \multicolumn{2}{>{\hsize=\dimexpr2\hsize+2\tabcolsep+\arrayrulewidth\relax}X}{\quad 3. The delicate balance of our planet's systems, from the tides that shape our coastlines to the complex interactions between plant and animal species, highlights the need for greater understanding and protection of these systems in the face of ongoing environmental changes.} \\
    \multicolumn{2}{>{\hsize=\dimexpr2\hsize+2\tabcolsep+\arrayrulewidth\relax}X}{\textbf{Revised Situations:}} \\
    \multicolumn{2}{>{\hsize=\dimexpr2\hsize+2\tabcolsep+\arrayrulewidth\relax}X}{\quad 1. Although our evolving dynamic planet has survived sea level changes of hundreds of metres, the rapid rise of sea levels caused by climate change has led to the destruction of many coastal cities and ecosystems, demonstrating the vulnerability of biological, geological, and planetary systems.} \\
    \multicolumn{2}{>{\hsize=\dimexpr2\hsize+2\tabcolsep+\arrayrulewidth\relax}X}{\quad 2. The geological history of our planet is marked by numerous catastrophic events, such as massive volcanic eruptions and asteroid impacts, which have had a significant impact on the evolution of life on Earth. However, our evolving dynamic planet has survived sea level changes of hundreds of metres.} \\
    \multicolumn{2}{>{\hsize=\dimexpr2\hsize+2\tabcolsep+\arrayrulewidth\relax}X}{\quad 3. Our evolving dynamic planet has survived sea level changes of hundreds of metres, but the delicate balance of our planet's systems, from the tides that shape our coastlines to the complex interactions between plant and animal species, highlights the need for greater understanding and protection of these systems in the face of ongoing environmental changes.} \\
    \multicolumn{2}{>{\hsize=\dimexpr2\hsize+2\tabcolsep+\arrayrulewidth\relax}X}{\textbf{Prediction:} Insufficient}\\
    \bottomrule
    \end{tabularx}
    \caption{An example of the detailed reasoning process of \textsc{Casa} (\textsc{Tulu}) on Climate.}
    \label{table-case-climate}
\end{table*}

\subsection{Experimental Setup}
We experiment with two instruction-tuned models: \textsc{Tulu-7b}~\citep{wang2023far} and \textsc{Llama-2-7b-chat}~\citep{touvron2023llama2} as the base models of our framework. \textsc{Tulu-7b} is finetuned on \textsc{Llama-7b}~\citep{touvron2023llama} with an aggregation of instruction tuning datasets, achieving great performance across benchmarks. We do not choose the prevalent GPT models like GPT-4 because portions of BIG-bench data were mixed into its training set~\citep{openai2023gpt4}.

We use an off-the-shelf negator~\citep{anschutz2023not} based on syntactic rules to generate $\neg \texttt{Premise}$ and $\neg \texttt{Conclusion}$, and use BART-NLI\footnote{\url{https://huggingface.co/facebook/bart-large-mnli}} as the NLI model used in probability estimation. We sample $n=3$ units for each argument, and make the final decision with a majority vote. More implementation details are in Appendix~\ref{sec-appendix-implementation}.

There is no existing zero-shot argument sufficiency assessment model to our knowledge, so we build several non-trivial baselines for comparison:

\vspace{1mm}
\noindent\textbf{Zero-shot Prompting.} We probe the base models \textsc{Tulu} and \textsc{Llama2} with four prompt forms (two forms provided by BIG-bench-LFD, and two forms written by ourselves), and report the best performance. The detailed prompts are in Appendix~\ref{sec-appendix-baseline}.

\vspace{1mm}
\noindent\textbf{Perplexity.} Motivated by~\citet{zhang2022rock}, we compute perplexity scores for the base models as another zero-shot baseline. For each argument, we compare the perplexity score of $\texttt{Premise} || \texttt{Conclusion}$ and $\texttt{Premise} || \neg \texttt{Conclusion}$, and regard the one with lower perplexity score as the model prediction. Here $||$ indicates text concatenation.

\vspace{1mm}
\noindent\textbf{NLI Models.} We directly use two NLI models, RoBERTa-NLI~\citep{reimers-gurevych-2019-sentence} and BART-NLI, to conduct the task. Specifically, NLI models are asked to predict if the premises support or contradict the conclusion.

\vspace{1mm}
\noindent\textbf{One-shot Prompting.} Besides the zero-shot baselines, we add one-shot prompting into comparison, to help LLMs better understand the task instruction. We use the prompt form that performs the best in zero-shot prompting, and randomly select one example from the datasets. We test models three times with different examples and report the average performance.

\subsection{Results}
\begin{figure*}[th]
    \centering
    \includegraphics[width=0.9\textwidth]{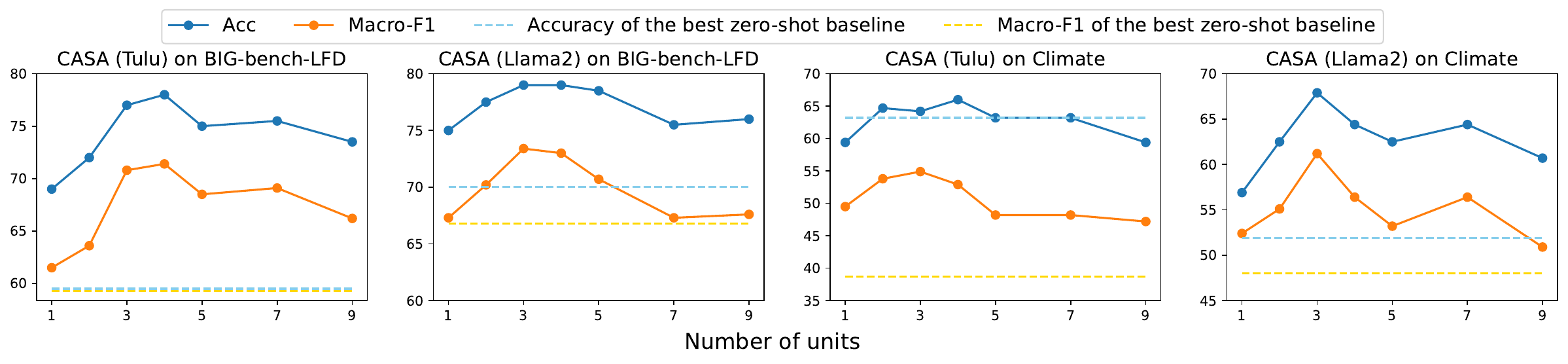}
    \caption{Hyperparameter analysis on the number of units. \textsc{Casa} consistently outperforms baselines on macro-F1.} 
    \label{fig-hyper}
\end{figure*}
\noindent\textbf{\textsc{Casa} vs. Baselines.}
Table~\ref{table-main} reports the automatic evaluation results. \textsc{Casa} significantly outperforms all the corresponding zero-shot baselines with significance level $\alpha = 0.02$, and also surpasses the one-shot baselines. This demonstrates the effectiveness of our overall framework. 
The performance on Climate is inferior to on BIG-bench-LFD for all models, because Climate requires specific domain knowledge and more rigorous reasoning, as some fallacies are less apparent. \textsc{Casa} shares the same domain knowledge with the baselines as they use the same base models, but the causality-driven framework equips it with better reasoning performance. We demonstrate a case in Section~\ref{sec-case}.

\vspace{1mm}
\noindent\textbf{Step-wise Evaluation.}
To examine whether LLMs work as we expect in each step of \textsc{Casa}, we conduct step-wise human evaluation. We ask human annotators to rate three aspects individually: 
1) In the claim extraction step, do LLMs extract the correct premises and conclusion from the argument? 
2) In the context sampling step, are the contexts generated by LLMs consistent with $\neg \texttt{Premise}$ and $\neg \texttt{Conclusion}$? 
3) In the revision step, are the revised situations consistent with the $\texttt{Premise}$?
We sample 100 instances from BIG-bench-LFD and recruit three annotators to answer each question. We report the majority vote results, and the inter-annotator agreement is 84\%.

Table~\ref{table-stepwise} shows the step-wise evaluation results. The accuracy of all aspects is above 90\%, exhibiting that LLMs are capable of generating textual data that conform to certain conditions, and making interventions on situations in the form of natural language. 
We provide the annotation templates and error analysis in Appendix~\ref{sec-appendix-stepwise}. 
\section{Analysis}
\begin{figure*}[th]
    \centering
    \includegraphics[width=0.9\textwidth]{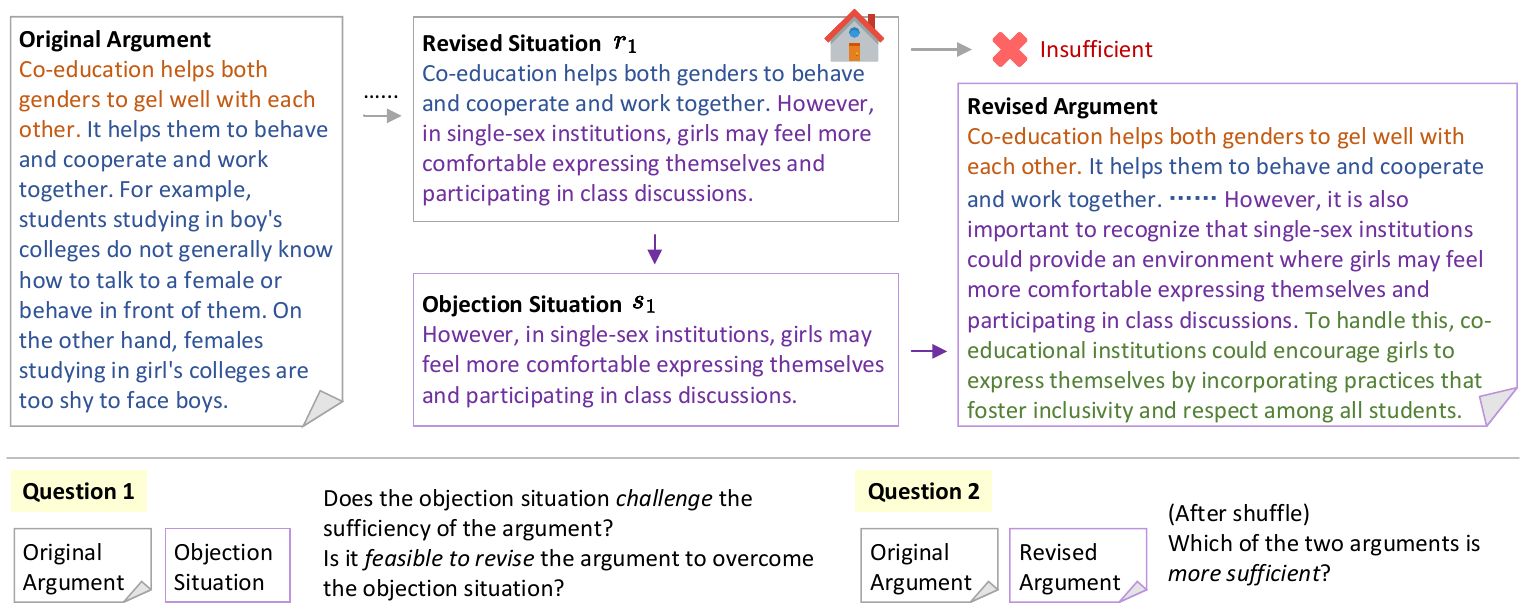}
    \caption{The process of providing writing assistance with \textsc{Casa} (top) and conducting human evaluations for the assistance effectiveness (bottom).} 
    \label{fig-application}
\end{figure*}

\begin{table*}[ht]
    \centering
    \small
    \begin{subtable}[h]{0.48\textwidth}
        \centering
        \scalebox{0.9}{
        \begin{tabular}{lcc}
        \toprule
        \textbf{Model} & \textbf{Rationality} & \textbf{Feasibility} \\
        \midrule
        Prompting (\textsc{Tulu}) & 84\% & 60\% \\
        Prompting (\textsc{Llama}2) & 90\% & 67\% \\
        \textsc{Casa} (\textsc{Tulu}) & 90\% & 76\% \\
        \textsc{Casa} (\textsc{Llama}2) & \textbf{92\%} & \textbf{81\%}\\
        \bottomrule
        \end{tabular}
        }
        \caption{Question 1: Is \textsc{Casa} capable of generating rational and feasible objection situations to the essays?}
        \label{table-q1}
    \end{subtable}
    \hspace{6pt}
    \begin{subtable}[h]{0.48\textwidth}
        \centering
        \scalebox{0.9}{
        \begin{tabular}{lccc}
        \toprule
        & \textbf{Revised is} & & \textbf{Original is} \\
        \multirow{-2}{*}{\textbf{Model}} & \textbf{Better} & \multirow{-2}{*}{\textbf{Tie}} & \textbf{Better} \\
        \midrule
        Prompting (\textsc{Tulu}) & \textbf{41\%} & 38\% & 21\% \\
        Prompting (\textsc{Llama}2) & \textbf{51\%} & 23\% & 26\% \\
        \textsc{Casa} (\textsc{Tulu}) & \textbf{40\%} & 43\% & 17\% \\
        \textsc{Casa} (\textsc{Llama}2) & \textbf{58\%} & 35\% & 7\% \\
        \bottomrule
        \end{tabular}
        }
        \caption{Question 2: Will revising based on the generated objection situations improve the sufficiency of the essays?}
        \label{table-q2}
    \end{subtable}
    \caption{Human evaluation results of using \textsc{Casa} to provide writing suggestions.}
    \label{table-suggestion}
\end{table*}

\subsection{Ablation Study}
To further investigate the effectiveness of \textsc{Casa} components, we study several variants of \textsc{Casa}:

\noindent\textbf{w/o Intervention.} In this ablation, we simply estimate $P(Y=1|X=1)$ without intervention. Concretely, we sample contexts for $\texttt{Premise}$, and estimate the probability of $\texttt{Conclusion}$ based on contexts and $\texttt{Premise}$.

\noindent\textbf{w/o Condition on $X = 0$.} This variant estimates $P(Y(X=1)=1|Y=0)$, where the term $X=0$ is removed from the original $PS$ definition. In the context sampling step, we ask LLMs to generate contexts only consistent with $\neg \texttt{Conclusion}$, and other steps are kept the same.

\noindent\textbf{w/o Condition on $Y = 0$.} This variant estimates $P(Y(X=1)=1|X=0)$, and LLMs are asked to generate contexts consistent with $\neg \texttt{Premise}$.

\noindent\textbf{Intervention: Concatenation.} We study if the intervention step can be replaced by simply concatenating the context with $\texttt{Premise}$. In this setting, the mediator $Z$ in the context remains unchanged. 

Table~\ref{table-ablation-bigbench} shows the ablation results on BIG-bench-LFD, and results on Climate are in Appendix Table~\ref{table-ablation-climate}. All the ablations lead to performance drops on macro-F1, indicating that the original probability of sufficiency definition is not only of theory value, but also of practical value.

\subsection{Hyperparameter Study}
To study the performance sensitivity of \textsc{Casa} to the number of units $n$, we vary $n$ from $1$ to $9$ for \textsc{Casa (Tulu)} and \textsc{Casa (Llama2)}, and exhibit the performance in Figure~\ref{fig-hyper}. 

\textsc{Casa} consistently outperforms zero-shot baseline models on macro-F1 regardless of the number of units, proving its robustness towards the hyperparameter. The performance peak is around $3$ in all settings. When we sample too few units, they hardly encompass a wide variety of situations. On the other hand, when we sample many contexts at once, the quality of the contexts goes down and each context tends to be shorter.

\subsection{Case Study}
\label{sec-case}
We demonstrate an example of the reasoning process of \textsc{Casa} on Climate in Table~\ref{table-case-climate}, and an example on BIG-bench-LFD in Appendix Table~\ref{table-case-bigbench}. In both cases, \textsc{Casa} is able to detect the insufficiency in the argument, while directly prompting the same base model fails to find the fallacy. Specifically, in Table~\ref{table-case-climate}, \textsc{Casa} generates several evidences supporting that \textit{biological systems are vulnerable}, like \textit{the sea level rise leads to the destruction of coastal cities} and \textit{volcanic eruptions impact the evolution of life}. These evidences do not contradict with the premise that \textit{our planet has survived sea level changes}. Therefore, they are kept in the revised situations, and make the conclusion \textit{biological systems are extremely robust} no longer supported.

\section{Application: Writing Assistance}
\label{sec-application}
We apply \textsc{Casa} to a realistic scenario: providing writing suggestions for essays written by students. If \textsc{Casa} identifies that an argument in an essay is insufficient, we extract explainable reasons from \textsc{Casa}'s reasoning process, and provide them as suggestions for revision.

Specifically, we generate \textit{objection situations} (situations that challenge the sufficiency of the argument) out of intervened situations $R$ that contradict the $\texttt{Conclusion}$, by removing the $\texttt{Premise}$ from $R$. As shown in the example of Figure~\ref{fig-application}, the revised situation $r_1$ is converted to the objection situation $s_1$ by removing the premise \textit{co-education helps both genders to behave and cooperate and work together.} The removal is automatically done by detecting if each sentence in the revised situation entails one premise event.

We investigate 
 \textit{(Q1):} whether \textsc{Casa} is capable of generating rational and feasible objection situations to the essays, and \textit{(Q2):} whether revising based on the generated objection situations improves the sufficiency of the essays.

\vspace{1mm}
\noindent\textbf{Dataset.} The Argument-Annotated Essays (AAE,~\citet{stab2017parsing}) dataset contains 402 argumentative essays and 1,029 extracted arguments. \citet{stab-gurevych-2017-recognizing} use AAE to annotate argument sufficiency. They find that some arguments in this dataset suffer from the insufficiency problem, but the criteria for insufficiency are vague among annotators.\footnote{Although the overall agreement between annotators is 0.77 measured with Krippendorff's $\alpha$, on instances at least one annotator labels as insufficient (32\% of all instances), the agreement is only 0.14.} Therefore, we only use the corpus, but not the annotations.

We randomly sample 100 arguments from AAE, and assess their sufficiency with \textsc{Casa}. For the arguments \textsc{Casa} finds insufficient, we randomly transform one revised situation into the objection situation. We recruit three annotators to answer each question, and the annotation templates are in Appendix~\ref{sec-appendix-application}.

\vspace{1mm}
\noindent\textbf{Objection Quality.} Given an argument and an objection generated by \textsc{Casa}, we ask annotators to evaluate whether the objection challenges the sufficiency of the argument (rationality), and whether it is feasible to revise the argument to overcome the objection (feasibility). We also build a baseline of directly prompting LLMs to generate an objection situation if they identify the argument as insufficient. The prompt is in Appendix Table~\ref{table-prompt-objection}.

As shown in Table~\ref{table-q1}, objection situations generated by \textsc{Casa} are more rational and feasible than directly prompting the base models. The gap in feasibility is larger, as LLMs are likely to generate abstract objections when prompting, like \textit{the argument does not consider the potential challenges that may arise when students from different gender backgrounds interact}, while \textsc{Casa} provides more practical objections which are easier to address.

\vspace{1mm}
\noindent\textbf{Effect of the Revision.} To evaluate the sufficiency of revised arguments, we first use GPT-4~\citep{openai2023gpt4} to revise the arguments based on the objection situations generated by the prompting method and \textsc{Casa}. This is to simulate a human revision process, since asking human annotators to complete this task is expensive and hard to control the quality. We further recruit annotators to check the revised arguments to ensure the revision quality. The revision prompt and quality check are described in Appendix~\ref{sec-appendix-application}. 

We ask annotators to compare the sufficiency of the original and revised arguments. As shown in Table~\ref{table-q2}, in both methods tested, the Revised is Better proportion supersedes the Original is Better proportion, emphasizing an improvement in writing sufficiency.
On the other hand, with the same base model, CASA obtains a higher Revised - Original ratio (the Revised is Better proportion minus the Original is Better proportion) compared to the prompting method. This suggests that, even if we do not consider the difficulty of revision, CASA helps more in the revision process.
\section{Related Work}
\noindent\textbf{Argument Sufficiency Assessment.} Previous works on argument sufficiency assessment mainly use standard inference models. \citet{wachsmuth-werner-2020-intrinsic} train support vector machines (SVM) on text features;~\citet{stab-gurevych-2017-recognizing} use convolutional neural networks (CNN) to recognize insufficient arguments; and~\citet{saveleva-etal-2021-graph} employ graph neural networks (GNN) to better understand argument structures. However, they all suffer from the subjective criteria of annotation~\citep{rach-etal-2020-evaluation,wachsmuth-etal-2017-computational}. Additionally, they simply treat the task as a normal classification task without considering the nature of sufficiency. \citet{gurcke-etal-2021-assessing} model the relation between premises and conclusion of a sufficient argument, but the relation is based on their personal hypotheses, whereas \textsc{Casa} possesses a clear theoretical foundation.

\vspace{1mm}
\noindent\textbf{Writing Assistance.} There are also previous works trying to provide writing assistance to human-written articles. \citet{hanawa-etal-2021-exploring, zhang-etal-2022-automatic-comment, wang-etal-2023-smart} focus on polishing the form of the writing, like grammatical correctness, word choices and rhetorical methods. \citet{wambsganss-niklaus-2022-modeling} try to provide feedback on argument content, but the feedback is mostly given as scores on predefined dimensions. In contrast, \textsc{Casa} provides feedback in the form of objection situations, which makes the revision easier. \citet{skitalinskaya-wachsmuth-2023-revise} discuss how to identify argumentative claims that need further revision, which is complementary to our work. 
\section{Conclusion}
We propose \textsc{Casa}, a zero-shot argument sufficiency assessment framework driven by the causal concept of sufficiency. In the absence of observational data and intervention data, we sample contexts and make interventions with LLMs. \textsc{Casa} is capable of identifying insufficient arguments on two logical fallacy detection datasets, and providing writing suggestions to further improve the sufficiency of human-written arguments.

\section*{Acknowledgments}
We thank Xueqing Wu, Fan Yin, Ashima Suvarna, Da Yin, and other UCLA-NLP lab members for their constructive comments. We thank the anonymous reviewers for their helpful discussions and suggestions. This research was supported in part by NSF \#2331966.

\section*{Limitations}
\paragraph{Choices in Model Design.} We tried to explore diverse decoding methods when sampling contexts, as they may generate a large number of diverse contexts without impairing the quality of each context. However, we find that in our scenario, current diverse decoding methods can hardly generate high-quality contexts that are diverse in content, so we tend to instruct LLMs to generate multiple contexts in one run. 

The goal of the revision under intervention step can be viewed as a counterfactual reasoning task, so we also explore zero-shot counterfactual reasoning models for this step, but they are either slow in inference or ineffective in the generation quality when conducting the revision task. Therefore, we prompt LLMs to complete the step. We are willing to switch to new methods for these steps if more powerful diverse decoding/counterfactual reasoning methods are released.

\paragraph{Data Scope.} Evaluating model performances on the argument sufficiency assessment task is difficult, due to the aforementioned subjective annotation criteria. Although we try our hardest to find automatic evaluation datasets, the two datasets we use are still of limited scope. To make up for this, we calculate the significance level and conduct diverse analyses.
\section*{Ethics Statement}
Our framework can be applied to educational scenarios like student essay assessment and comment generation. However, as we cannot ensure the model prediction is correct, it must be used with manual check. 
Moreover, as our framework is based on existing LLMs, its generations may inherit the bias of LLMs. Therefore it should be used under human supervision. 

For human evaluations, we recruit annotators from Amazon Mechanical Turk, and all annotators are fairly paid more than \$10 USD per hour (it varies depending on the time spent on HITs), which is higher than the national minimum wage where the annotators are recruited.

\bibliography{anthology,custom}

\appendix
\section{Appendix}
\label{sec-appendix}
\subsection{Implementation Details}
\label{sec-appendix-implementation}
\begin{table*}[ht]
    \centering
    \small
    \renewcommand{\arraystretch}{1.2}
    \begin{tabularx}{\textwidth}{X}
    \toprule
    \textbf{\#\#\# Instruction:} \\
    Determine which part of the text is the conclusion. \\
    Output the number of the conclusion part first, and give an explanation.\\
    Format:\\
    Conclusion: [number]\\
    Explanation: ...\\
    \textbf{\#\#\# Input:}\\
    You shouldn't trust Donald's views about politics. He's an alcoholic. \\
    Choices: \\
    1. You shouldn't trust Donald's views about politics. \\
    2. He's an alcoholic. \\
    \textbf{\#\#\# Response:}\\
    \bottomrule
    \end{tabularx}
    \caption{Example prompt for claim extraction.}
    \label{table-prompt-extraction}
\end{table*}

\begin{table*}[ht]
    \centering
    \small
    \renewcommand{\arraystretch}{1.2}
    \begin{subtable}[h]{\textwidth}
    \begin{tabularx}{\textwidth}{X}
    \toprule
    \textbf{\#\#\# Instruction:} \\
    Generate 3 detailed contexts. Each context is consistent with both the premise and the conclusion. Each context is in one line. \\
    \textbf{\#\#\# Input:}\\
    Premise: He isn't an alcoholic.\\
    Conclusion: You should trust Donald's views about politics.\\
    \textbf{\#\#\# Response:}\\
    \bottomrule
    \end{tabularx}
    \caption{Single premise.}
    \end{subtable}
    \begin{subtable}[h]{\textwidth}
    \begin{tabularx}{\textwidth}{X}
    \toprule
    \textbf{\#\#\# Instruction:} \\
    Generate 3 detailed contexts. Each context contains ``Positive things are good.'' Each context is consistent with both the premise and the conclusion. Each context is in one line. \\
    \textbf{\#\#\# Input:}\\
    Premise: My drug test wasn't positive.\\
    Conclusion: My test result wasn't good.\\
    \textbf{\#\#\# Response:}\\
    \bottomrule
    \end{tabularx}
    \caption{Multiple premises. The argument is ``My drug test was positive, and positive things are good. So my test result was good.''}
    \end{subtable}
    \caption{Example prompts for context sampling.}
    \label{table-prompt-sample}
\end{table*}

\begin{table*}[ht]
    \centering
    \small
    \renewcommand{\arraystretch}{1.2}
    \begin{tabularx}{\textwidth}{X}
    \toprule
    \textbf{\#\#\# Instruction:} \\
    Revise the text to contain the provided statement.\\
    \textbf{\#\#\# Input:}\\
    Text: Donald's political views are based on his own personal experiences and observations, which have been shaped by his sober perspective. \\
    Statement: He's an alcoholic. \\
    \textbf{\#\#\# Response:}\\
    \bottomrule
    \end{tabularx}
    \caption{Example prompt for revision under intervention.}
    \label{table-prompt-revision}
\end{table*}
We demonstrate examples of the prompts we used in \textsc{Casa}. Table~\ref{table-prompt-extraction} showcases the prompt for claim extraction. The argument is segmented using rules of punctuation marks and conjunction words. Specifically, we first check if the argument can be split with punctuation marks like periods and semicolons. If not, we further split it with conjunction words like \textit{because} and \textit{so}.

Table~\ref{table-prompt-sample} shows the prompts for context sampling. The premise and conclusion in the prompts are $\neg \texttt{Premise}$ and $\neg \texttt{Conclusion}$ of the argument. When the argument contains multiple premises, we check one premise each time, and ask LLMs to generate contexts containing other premises, like \textit{positive things are good} in the example.

Table~\ref{table-prompt-revision} shows the prompt for revision under intervention. The statement here is the premise of the argument, and the text is a context generated in the last step.

When probing \textsc{Tulu}, we add the prefix ``<|user|>'' and suffix ``<|assistant|>'' to the prompts, aligning with the original instruction tuning format~\citep{wang2023far}.

Experiments are conducted on NVIDIA RTX A6000 GPUs. It takes \textsc{Casa} two minutes to determine the sufficiency of one argument on average.

\subsection{Details of Prompting Baselines}
\label{sec-appendix-baseline}
\begin{table*}[ht]
    \centering
    \small
    \renewcommand{\arraystretch}{1.2}
    \begin{tabularx}{\textwidth}{lX}
    \toprule
    \textbf{Prompt 1} & \textbf{\#\#\# Instruction:}\\
    & This AI is identifying whether statements contain fallacies. The AI responds with `Valid' or `Invalid' as appropriate.\\
    & \textbf{\#\#\# Input:}\\
    & You shouldn't trust Donald's views about politics. He's an alcoholic.\\
    & \textbf{\#\#\# Response:}\\
    \textbf{Prompt 2} & \textbf{\#\#\# Instruction:}\\
    & Do you think the following argument is `Valid' or `Invalid'?\\
    & \textbf{\#\#\# Input:}\\
    & You shouldn't trust Donald's views about politics. He's an alcoholic.\\
    & \textbf{\#\#\# Response:}\\
    \textbf{Prompt 3} & \textbf{\#\#\# Instruction:}\\
    & This AI is identifying whether arguments are sufficient. The AI responds with `Sufficient' or `Insufficient' as appropriate.\\
    & \textbf{\#\#\# Input:}\\
    & You shouldn't trust Donald's views about politics. He's an alcoholic.\\
    & \textbf{\#\#\# Response:}\\
    \textbf{Prompt 4} & \textbf{\#\#\# Instruction:}\\
    & Do you think the following argument is `Sufficient' or `Insufficient'?\\
    & \textbf{\#\#\# Input:}\\
    & You shouldn't trust Donald's views about politics. He's an alcoholic.\\
    & \textbf{\#\#\# Response:}\\
    \bottomrule
    \end{tabularx}
    \caption{Prompts explored for the zero-shot prompting baselines.}
    \label{table-prompt-baseline}
\end{table*}
BIG-bench-LFD provides two prompt forms, and we follow these prompts in our evaluation. Moreover, we modify the prompts to explicitly ask LLMs to predict ``sufficient'' or ``insufficient''. These lead to the four prompts shown in Table~\ref{table-prompt-baseline}. 

When probing \textsc{Llama2}, we further try two options: 1) directly use the prompts, and 2) wrap them up with ``<s>[INST] <<SYS>>\{\{system\_prompt\}\}<</SYS>>\{\{our\_prompt\}\} [/INST]''.\footnote{The system prompt is shown in \url{https://github.com/huggingface/blog/blob/main/llama2.md}} The second option is suggested by the authors of \textsc{Llama2}, but it sometimes refuses to answer questions. When probing \textsc{Tulu}, we add the prefix ``<|user|>'' and suffix ``<|assistant|>'' to the prompts. 
We evaluate LLMs with all the prompts, and report the performance of the prompt with the highest macro-F1 on the test set.

\subsection{Step-wise Evaluation}
\label{sec-appendix-stepwise}
\begin{figure*}[th]
    \centering
    \includegraphics[width=\textwidth]{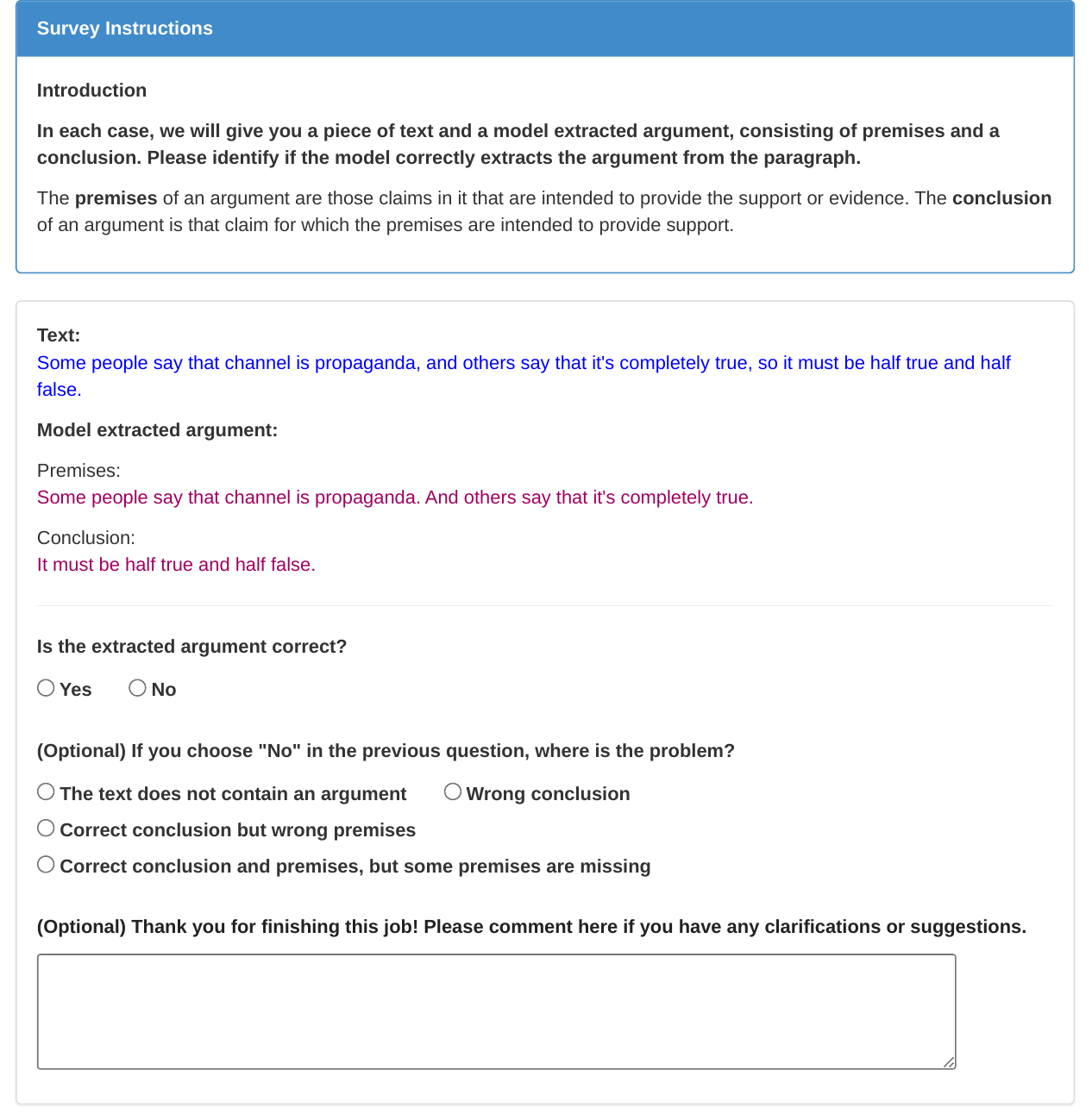}
    \caption{Annotation template for step-wise evaluation: correctness of claim extraction.} 
    \label{fig-template-extraction}
\end{figure*}
\begin{figure*}[th]
    \centering
    \includegraphics[width=\textwidth]{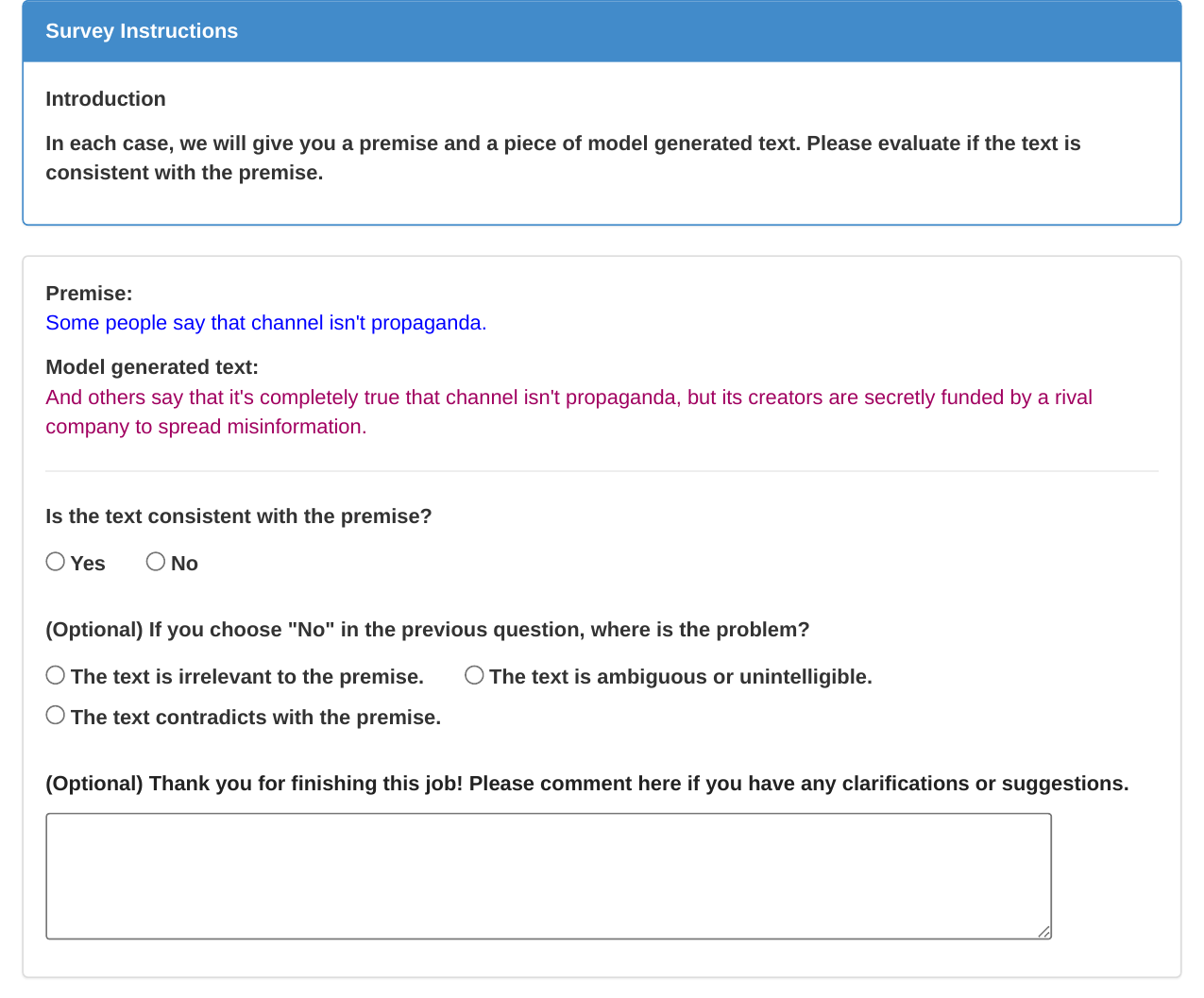}
    \caption{Annotation template for step-wise evaluation: consistency of the generated context with $\neg \texttt{Premise}$.} 
    \label{fig-template-consistency}
\end{figure*}
We demonstrate the annotation templates of step-wise evaluation in Figure~\ref{fig-template-extraction} for claim extraction, and Figure~\ref{fig-template-consistency} for consistency between the generated context and $\neg \texttt{Premise}$. Other consistency evaluations are similar to Figure~\ref{fig-template-consistency}.

We also ask annotators to choose the error reason if they feel the extracted annotation is wrong, or the generated text is inconsistent with the given statement. For claim extraction, all four error types shown in Figure~\ref{fig-template-extraction} exist, while \textit{wrong conclusion} occurs more often, in 40\% of the incorrect cases.
For consistency, the generated text contradicts with the given statement in about 50\% of the errors; the text is irrelevant to the given statement in 40\% errors; and the text is ambiguous or unintelligible in 10\% errors.

Additionally, we estimate whether the revised situations preserve the original content with BLEU~\citep{papineni2002bleu}. The BLEU score between revised situations and originally generated contexts is 57\% and 53\% for \textsc{Casa} (\textsc{Tulu}) and \textsc{Casa} (\textsc{Llama2}) respectively, demonstrating high-level content preservation.

\subsection{Additional Results}
\label{sec-appendix-additional}
\begin{table*}[ht]
    \centering
    \small
    \begin{subtable}[h]{0.48\textwidth}
        \centering
        \scalebox{0.9}{
        \begin{tabular}{lcc}
        \toprule
        \textbf{Model} & \textbf{Acc} & \textbf{Macro-F1} \\
        \midrule
        \rowcolor[gray]{0.95} \textsc{Casa} (\textsc{Tulu}) & 64.2 & \textbf{54.9} \\
        \quad w/o Intervention & 65.1 & 52.2 \\
        \quad w/o Condition on $X=0$ & \textbf{66.0} & 50.0 \\
        \quad w/o Condition on $Y=0$ & 59.4 & 53.1 \\
        \quad Intervention: Concatenation & 62.3 & 47.7 \\
        \bottomrule
        \end{tabular}
        }
        \caption{Ablations for \textsc{Casa} (\textsc{Tulu}).}
    \end{subtable}
    \hspace{6pt}
    \begin{subtable}[h]{0.48\textwidth}
        \centering
        \scalebox{0.9}{
        \begin{tabular}{lcc}
        \toprule
        \textbf{Model} & \textbf{Acc} & \textbf{Macro-F1} \\
        \midrule
        \rowcolor[gray]{0.95} \textsc{Casa} (\textsc{Llama}2) & \textbf{67.9} & \textbf{61.2}\\
        \quad w/o Intervention & 57.5 & 50.1 \\
        \quad w/o Condition on $X=0$ & 60.4 & 45.0 \\
        \quad w/o Condition on $Y=0$ & 60.4 & 53.0 \\
        \quad Intervention: Concatenation & \textbf{67.9} & 59.7 \\
        \bottomrule
        \end{tabular}
        }
        \caption{Ablations for \textsc{Casa} (\textsc{Llama}2).}
    \end{subtable}
    \caption{Ablation results on Climate. Numbers are in percentages (\%).}
    \label{table-ablation-climate}
\end{table*}
\begin{table*}[ht]
    \centering
    \small
    \renewcommand{\arraystretch}{1.2}
    \begin{tabularx}{\textwidth}{XX}
    \toprule
    \multicolumn{2}{>{\hsize=\dimexpr2\hsize+2\tabcolsep+\arrayrulewidth\relax}X}{\textbf{Argument:} I can't understand Higgs Theorem, so it must be false.}\\
    \textbf{Correct Answer:} Insufficient & \textbf{Zero-shot Prompting (\textsc{Llama2}):} Sufficient \\
    \rowcolor[gray]{0.95} \multicolumn{2}{>{\hsize=\dimexpr2\hsize+2\tabcolsep+\arrayrulewidth\relax}X}{\textsc{Casa} (\textsc{Llama2})}\\
    \multicolumn{2}{>{\hsize=\dimexpr2\hsize+2\tabcolsep+\arrayrulewidth\relax}X}{\textbf{Premise:} I can't understand Higgs Theorem.} \\
    \multicolumn{2}{>{\hsize=\dimexpr2\hsize+2\tabcolsep+\arrayrulewidth\relax}X}{\textbf{Conclusion:} It must be false.} \\
    \multicolumn{2}{>{\hsize=\dimexpr2\hsize+2\tabcolsep+\arrayrulewidth\relax}X}{\textbf{Sampled Contexts:}} \\
    \multicolumn{2}{>{\hsize=\dimexpr2\hsize+2\tabcolsep+\arrayrulewidth\relax}X}{\quad 1. As a physicist, I have spent years studying the Higgs boson and its role in the Standard Model of particle physics. I have a deep understanding of the mathematical concepts involved in the Higgs Theorem and have verified its validity through numerous experiments.} \\
    \multicolumn{2}{>{\hsize=\dimexpr2\hsize+2\tabcolsep+\arrayrulewidth\relax}X}{\quad 2. As a philosopher, I have explored the nature of truth and its relationship to mathematical concepts. I have come to understand that the Higgs Theorem is not just a mathematical construct, but a reflection of the fundamental laws of the universe. Its truth is not something that can be proven or disproven, but rather a reflection of the underlying reality.} \\
    \multicolumn{2}{>{\hsize=\dimexpr2\hsize+2\tabcolsep+\arrayrulewidth\relax}X}{\quad 3. As a detective, I have solved many complex cases involving mathematical puzzles and logical reasoning. I have a keen eye for detail and an ability to see patterns that others may miss. When I apply these skills to the Higgs Theorem, I am convinced that it must be true, as the evidence supports its validity.} \\
    \multicolumn{2}{>{\hsize=\dimexpr2\hsize+2\tabcolsep+\arrayrulewidth\relax}X}{\textbf{Revised Situations:}} \\
    \multicolumn{2}{>{\hsize=\dimexpr2\hsize+2\tabcolsep+\arrayrulewidth\relax}X}{\quad 1. As a physicist, I have spent years studying the Higgs boson and its role in the Standard Model of particle physics. But in fact, I can't understand Higgs Theorem without a deep understanding of the mathematical concepts involved in the Higgs Theorem and verifying its validity through numerous experiments.} \\
    \multicolumn{2}{>{\hsize=\dimexpr2\hsize+2\tabcolsep+\arrayrulewidth\relax}X}{\quad 2. As a philosopher, I have explored the nature of truth and its relationship to mathematical concepts. I have come to understand that the Higgs Theorem is not just a mathematical construct, but a reflection of the fundamental laws of the universe. In fact, I find it difficult to understand Higgs Theorem myself, as it seems to be a complex and abstract concept that defies easy comprehension. However, I believe that its truth lies beyond the realm of human understanding, and can only be appreciated through a deep contemplation of the fundamental laws of the universe.} \\
    \multicolumn{2}{>{\hsize=\dimexpr2\hsize+2\tabcolsep+\arrayrulewidth\relax}X}{\quad 3. As a detective, I have solved many complex cases involving mathematical puzzles and logical reasoning. I have a keen eye for detail and an ability to see patterns that others may miss. When I apply these skills to the Higgs Theorem, I am convinced that it must be true, as the evidence supports its validity. However, I must admit that I can't understand the Higgs Theorem myself, as it is far beyond my area of expertise.} \\
    \multicolumn{2}{>{\hsize=\dimexpr2\hsize+2\tabcolsep+\arrayrulewidth\relax}X}{\textbf{Prediction:} Insufficient}\\
    \bottomrule
    \end{tabularx}
    \caption{An example of the detailed reasoning process of \textsc{Casa} (\textsc{Llama2}) on BIG-bench-LFD.}
    \label{table-case-bigbench}
\end{table*}

Table~\ref{table-ablation-climate} shows the ablation results on Climate. Similar to the results on BIG-bench-LFD, all the ablations cause performance drops on macro-F1.

Table~\ref{table-case-bigbench} exhibits an example of the detailed reasoning process of \textsc{Casa(Llama2)} on BIG-bench-LFD.

\subsection{Details of the Writing Assistance Application}
\label{sec-appendix-application}
\begin{figure*}[th]
    \centering
    \includegraphics[width=\textwidth]{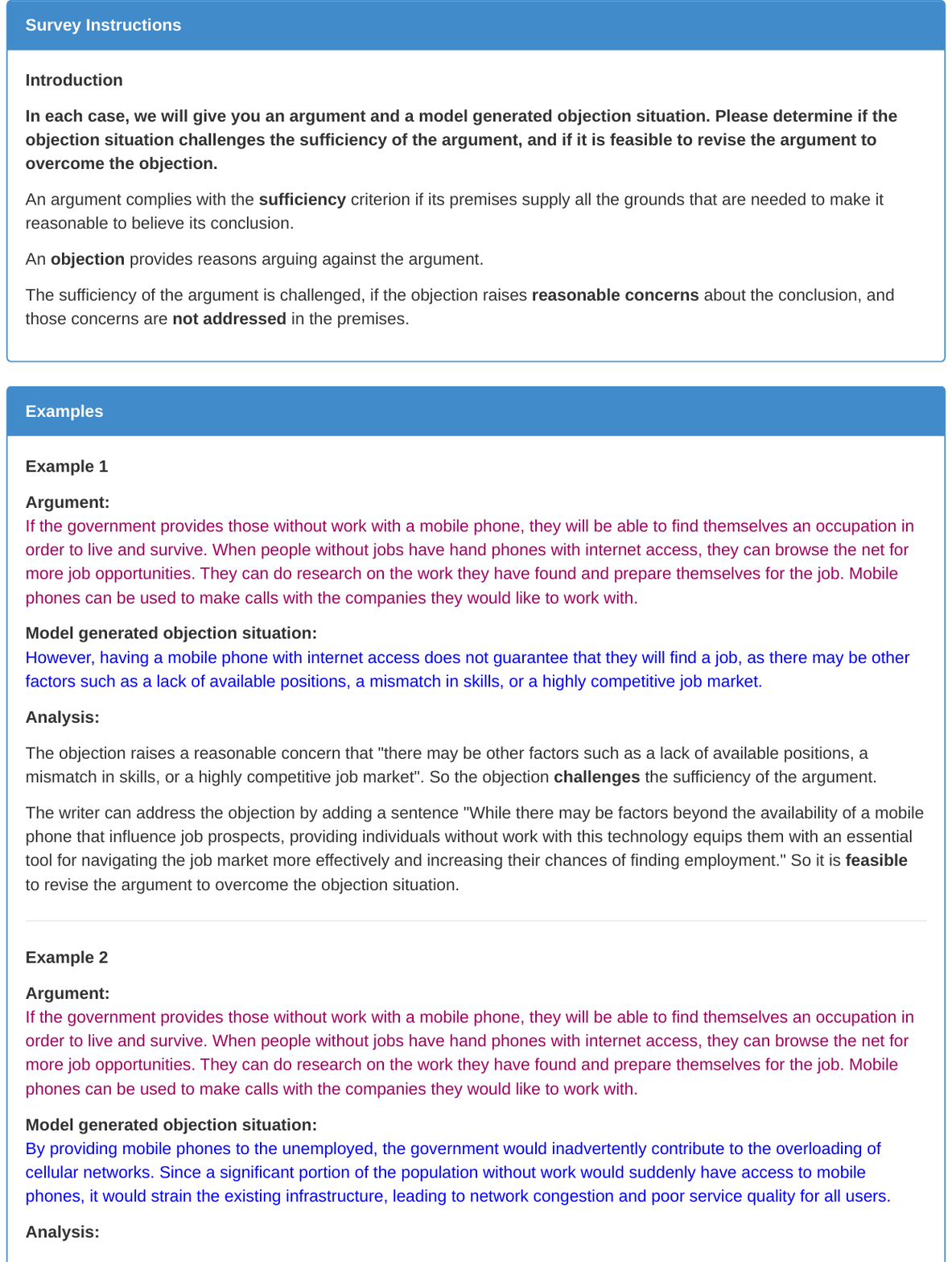}
    \caption{First half of the annotation template for evaluating the rationality and feasibility of objection situations. }
    \label{fig-template-objection}
\end{figure*}
\begin{figure*}[th]
    \ContinuedFloat
    \includegraphics[width=\textwidth]{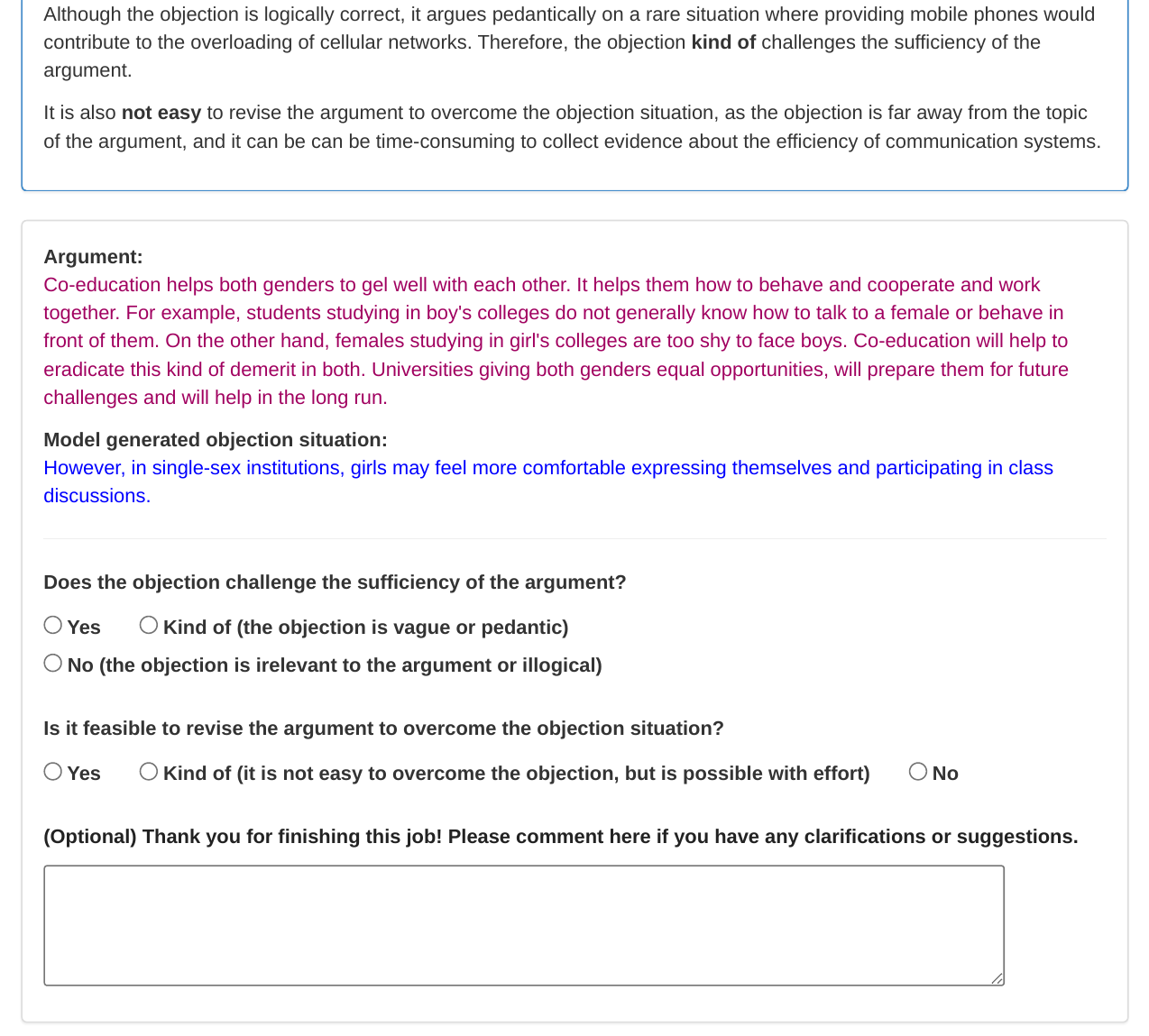}
    \caption{Second half of the annotation template for evaluating the rationality and feasibility of objection situations.} 
\end{figure*}

\begin{figure*}[th]
    \centering
    \includegraphics[width=\textwidth]{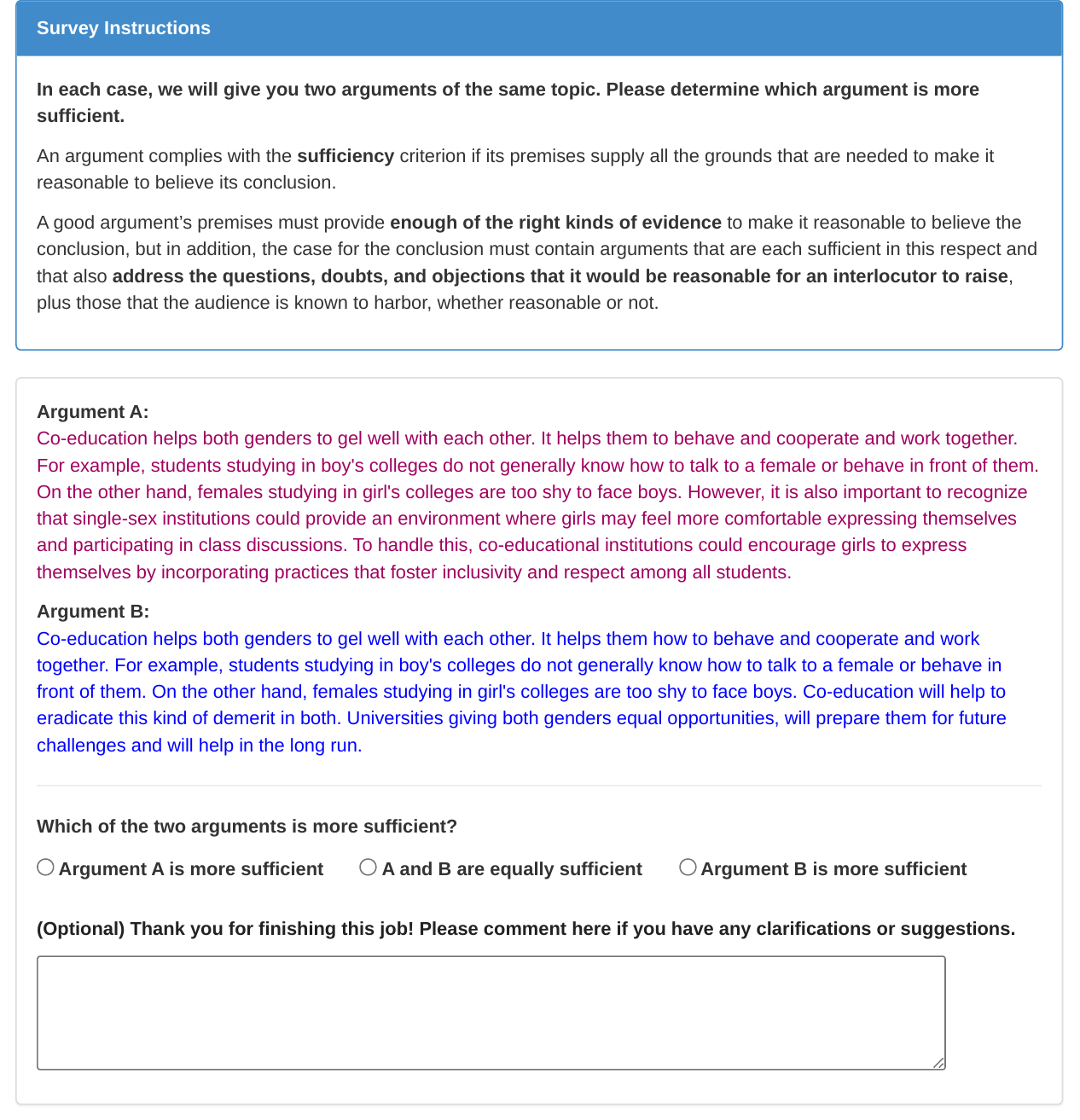}
    \caption{Annotation template for comparing the original and revised arguments.} 
    \label{fig-template-revision}
\end{figure*}

\begin{figure*}[th]
    \centering
    \includegraphics[width=\textwidth]{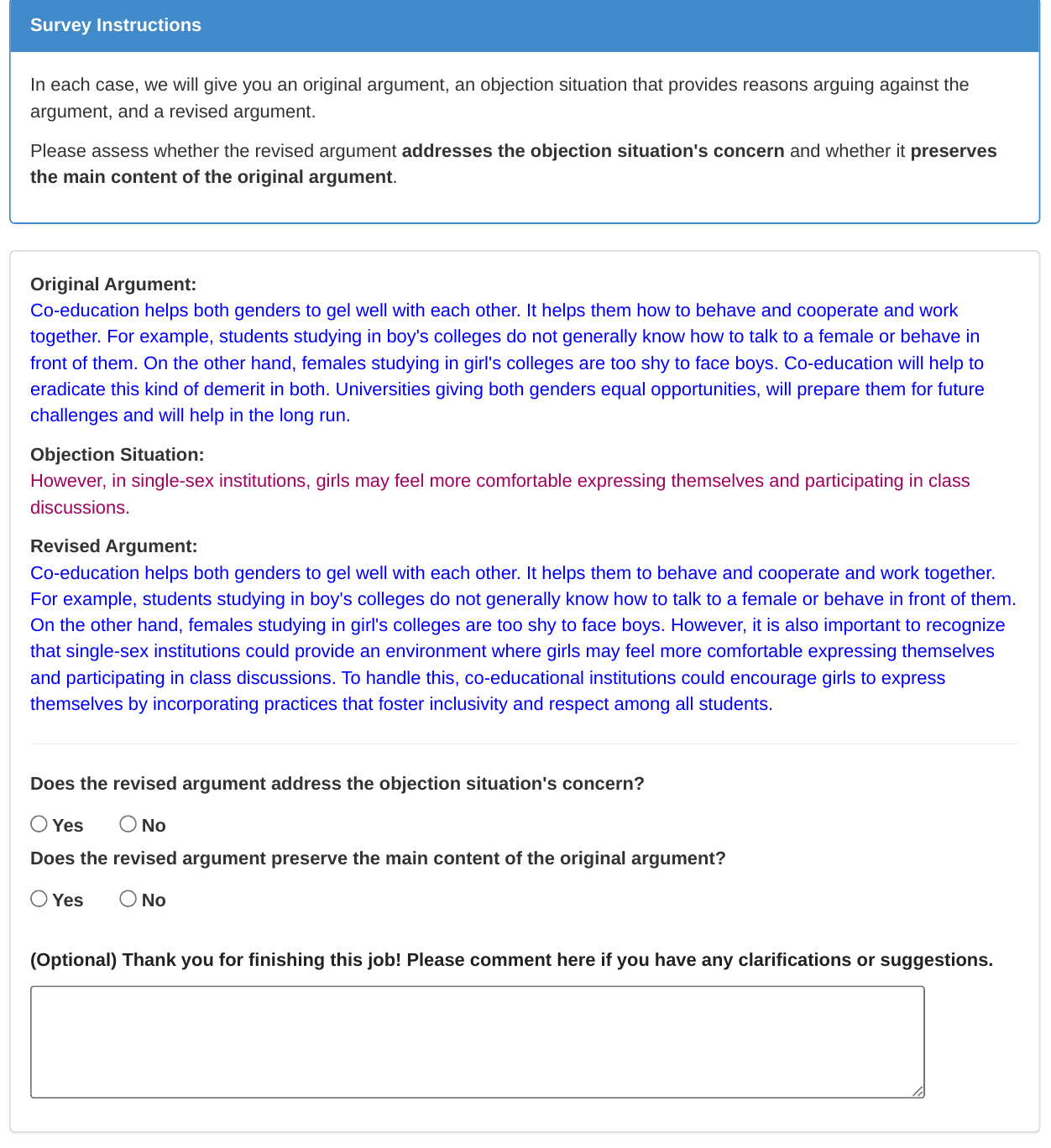}
    \caption{Annotation template for checking the revision quality of GPT-4.} 
    \label{fig-template-gpt-quality}
\end{figure*}

\begin{table*}[ht]
    \centering
    \small
    \renewcommand{\arraystretch}{1.2}
    \begin{tabularx}{\textwidth}{lX}
    \toprule
    \textbf{System} & You are a helpful and educated assistant. \\
    \textbf{User} & In each case, we will give you an argument and a model generated objection situation.\\
    & Your task is to revise the argument to address the concern raised in the objection situation. Please keep the conclusion and reasonable premises of the original argument unchanged. \\
    \textbf{User} & \textbf{Argument:} \\
    & Co-education helps both genders to gel well with each other. It helps them how to behave and cooperate and work together. For example, students studying in boy's colleges do not generally know how to talk to a female or behave in front of them. On the other hand, females studying in girl's colleges are too shy to face boys. Co-education will help to eradicate this kind of demerit in both. Universities giving both genders equal opportunities, will prepare them for future challenges and will help in the long run.\\
    & \textbf{Objection situation:}\\
    & However, in single-sex institutions, girls may feel more comfortable expressing themselves and participating in class discussions.\\
    & \textbf{Revised argument:}\\
    \bottomrule
    \end{tabularx}
    \caption{Example prompt for GPT-4 revision based on the objection situation.}
    \label{table-prompt-gpt4-revision}
\end{table*}
\begin{table*}[ht]
    \centering
    \small
    \renewcommand{\arraystretch}{1.2}
    \begin{tabularx}{\textwidth}{X}
    \toprule
    \textbf{\#\#\# Instruction:} \\
    This AI is identifying whether arguments are sufficient, capturing whether an argument's premises together make it rationally worthy of drawing its conclusion. The AI responds with 'Sufficient' or 'Insufficient' as appropriate. If the argument is insufficient, the AI also generates an objection situation to show the insufficiency.\\
    Format:\\
    Judgement: Sufficient or Insufficient\\
    Objection Situation (if insufficient): Describe a specific situation that challenges the sufficiency of the argument. Do not include any explanation.\\
    \textbf{\#\#\# Input:}\\
    In a positive point of view, when people without jobs have hand phones that have access to the Internet, they will be able to browse the net for more job opportunities. For example, they can surf the The Star Online's work section to find a job that is suitable for them. With the help of the net, they can also do more research on the work that they have found apart from looking up on how they can prepare themselves for the job. Not only that, the mobile phones can also be used to make calls with the companies in which they would like to work with. In short, if the government provides those without work with a mobile phone, they will be able to find themselves an occupation in order to live and survive.\\
    \textbf{\#\#\# Response:}\\
    Judgement: Insufficient\\Objection Situation: However, having a mobile phone with internet access does not guarantee that they will find a job, as there may be other factors such as a lack of available positions, a mismatch in skills, or a highly competitive job market.\\
    \textbf{\#\#\# Input:}\\
    Co-education helps both genders to gel well with each other. It helps them how to behave and cooperate and work together. For example, students studying in boy's colleges do not generally know how to talk to a female or behave in front of them. On the other hand, females studying in girl's colleges are too shy to face boys. Co-education will help to eradicate this kind of demerit in both. Universities giving both genders equal opportunities, will prepare them for future challenges and will help in the long run.\\
    \textbf{\#\#\# Response:}\\
    \bottomrule
    \end{tabularx}
    \caption{Example prompt for directly prompting LLMs to generate objection situations.}
    \label{table-prompt-objection}
\end{table*}

We demonstrate the annotation template for Question 1 \textit{(Is \textsc{Casa} capable of generating rational and feasible objection situations to the essays?)} in Figure~\ref{fig-template-objection}, and for Question 2 \textit{(Will revising based on the generated objection situations improve the sufficiency of the essays?)} in Figure~\ref{fig-template-revision}. The average inter-annotator agreement is 77\%. As the rationality and feasibility may be hard to understand, we provide two manually written examples and explanations for annotators. Although we design three options in Figure~\ref{fig-template-objection}, no question receives a majority vote of ``no'' in practice, so we only report the percentage of ``yes'' in Table~\ref{table-q1}.

The revision prompt for GPT-4 is shown in Table~\ref{table-prompt-gpt4-revision}. We use the version \texttt{gpt-4-0314} and temperature $0$. The prompt for the direct prompting baselines is in Table~\ref{table-prompt-objection}. To help LLMs better understand the instruction, we provide a human written example in the prompt. (In contrast, \textsc{Casa} is zero-shot.)

To evaluate the revision quality of GPT-4, we ask annotators to answer two questions as shown in Figure~\ref{fig-template-gpt-quality}: 1) Does the revised argument address the objection situation's concern? 2) Does the revised argument preserve the main content of the original argument? On 50 randomly sampled revisions from objections generated by \textsc{Casa (Tulu)} and \textsc{Casa (Llama2)}, 100\% of them address the objection, and 90\% of them preserve the main content of the original arguments.

\end{document}